\documentclass{article}
\usepackage{ijcai22}

\usepackage{times}
\usepackage{soul}
\usepackage{url}
\usepackage[hidelinks]{hyperref}
\usepackage[utf8]{inputenc}
\usepackage[small]{caption}
\usepackage{graphicx}
\usepackage{amsmath}
\usepackage{amsthm}
\usepackage{booktabs}
\usepackage{algorithm}
\usepackage{algorithmic}
\usepackage{subfigure}
\usepackage{multicol}
\usepackage{multirow}
\usepackage[switch]{lineno}
\usepackage{float}

\newcommand{\nosection}[1]{\vspace{2pt}\noindent\textbf{#1.}}
\title{Recent Advances in Reliable Deep Graph Learning: \\Inherent Noise, Distribution Shift, and Adversarial Attack}

\author{
Jintang Li$^{1}$\and
Bingzhe Wu$^{2}$\thanks{Corresponding author}\and
Chengbin Hou$^2$\and
Guoji Fu$^2$\and\\ 
Yatao Bian$^2$\and 
Liang Chen$^1$\and
Junzhou Huang$^3$\and
Zibin Zheng$^{1}$\\
\affiliations
$^1$Sun Yat-sen University\\
$^2$Tencent AI Lab\\
$^3$University of Texas at Arlington\\
\emails
lijt55@mail2.sysu.edu.cn,\{chenliang6,zhzibin\}@mail.sysu.edu.cn\\
\{bingzhewu,chengbinhou,guojifu,yataobian\}@tencent.com,\\
jzhuang@uta.edu
}

\begin{document}

\maketitle

\begin{abstract}
    Deep graph learning (DGL) has achieved remarkable progress in both business and scientific areas ranging from finance and e-commerce to drug and advanced material discovery. Despite the progress, applying  DGL  to real-world applications faces a series of reliability threats including inherent noise, distribution shift, and adversarial attacks. This survey aims to provide a comprehensive review of recent advances for improving the reliability of DGL algorithms against the above threats. In contrast to prior related surveys which mainly focus on adversarial attacks and defense, our survey covers more reliability-related aspects of DGL, i.e., inherent noise and distribution shift. Additionally, we discuss the relationships among above aspects and highlight some important issues to be explored in future research.
\end{abstract}

\section{Introduction}
Recent few years have seen deep graph learning (DGL) based on graph neural networks (GNNs) making remarkable progress in a variety of important areas,
ranging from business scenarios such as finance (e.g., fraud detection)~\cite{fraud_gnn} and e-commerce (e.g., recommendation system)~\cite{fedgnn_rec}, to scientific domains such as drug discovery and advanced material discovery~\cite{guo2021dockstream}.
DGL algorithms are impressive, but only when they work. In reality, they are largely unreliable.
Despite the progress, applying various DGL algorithms to real-world applications faces a series of reliability threats.
At a high level, we categorize these threats into three aspects, namely, \emph{inherent noise}, \emph{distribution shift}, and \emph{adversarial attack}.
Specifically, inherent noise refers to irreducible noises in graph structures, node attributes, and corresponding node/graph labels. Distribution shift refers to the shift between training and testing distribution which includes both domain generalization and sub-population shift.
Adversarial attack is a manipulative human action that aims to cause model misbehavior with carefully-designed patterns or perturbations on the original data, with typical examples including adversarial samples~\cite{DBLP:conf/kdd/ZugnerAG18} and backdoor triggers~\cite{DBLP:conf/uss/XiPJ021}.

\begin{table}[t]
    \centering
    \begin{tabular}{lc}
        \textbf{Threats}     & \textbf{Description}                                                                                     \\
        \midrule

        {Inherent Noise}     & $\mathcal{D}=(\mathbf{A}+\epsilon_a, \mathbf{X}+\epsilon_x, Y+\epsilon_y)$                               \\
        {Distribution shift} & $\text{P}_{\text{train}}(\mathcal{G}, Y) \neq \text{P}_{\text{test}}(\mathcal{G}, Y)$                    \\
        {Adversarial Attack} & $\hat{\mathcal{G}}= \arg \max_{\hat{\mathcal{G}} \approx \mathcal{G}} \mathcal{L}(\hat{\mathcal{G}}, Y)$ \\
        \bottomrule
    \end{tabular}
    \caption{Summary of recent advances in reliable deep graph learning. Oracle graph data $\mathcal{G} = (\mathbf{A}, \mathbf{X})$ with its adjacency matrix $\mathbf{A}$ and node features $\mathbf{X}$; $Y$ the labels of graphs/nodes; $\epsilon_a, \epsilon_x, \epsilon_y$ denote structure, attribute, and label noises respectively; $\mathcal{D}$ is the training dataset; $\mathcal{L}$ denotes the loss function and $\hat{\mathcal{G}}$ is the perturbed graph.}
    \label{tab:reliable_gnn_overview}
\end{table}

\begin{figure*}[t]
    \centering
    \includegraphics[width=\linewidth]{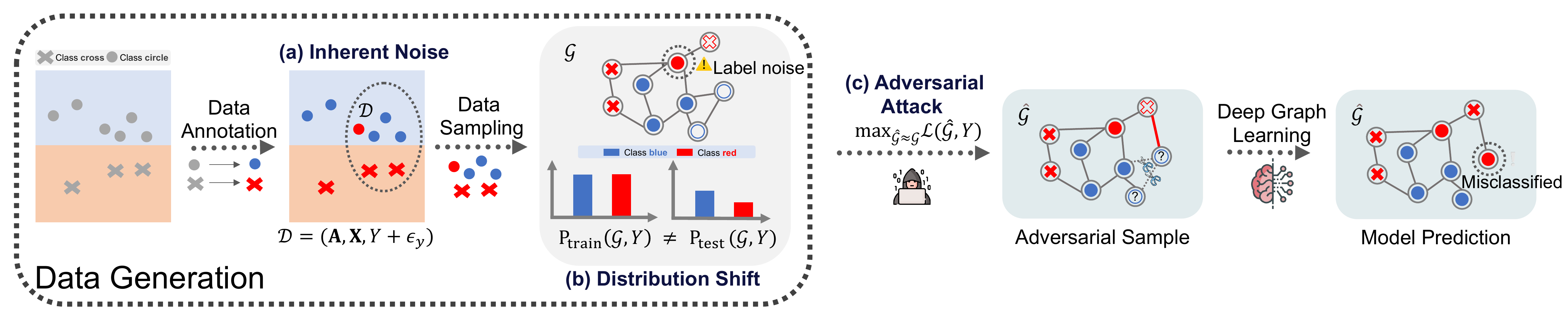}
    \caption{An illustrative example on deep graph learning against \textbf{(a)} inherent noise, \textbf{(b)} distribution shift, and \textbf{(c)} adversarial attack. From left to right: the inherent noise (i.e., label noise) is inevitably introduced during data annotation, where a red ``circle'' node is mislabeled with the ``cross'' class (marked red). As a result, the sampling bias leads to the discrepancy between training (color-filled nodes) and testing (color-unfilled nodes) datasets, introducing any possible kind of distribution shift.
        After the data generation process, adversarial attacks with a few edge manipulations on the graph are performed on the graph to mislead the model prediction (e.g., misclassification on a node).
    }
    \label{fig:framework}
\end{figure*}

Table~\ref{tab:reliable_gnn_overview} provides formal descriptions of each threat. Figure~\ref{fig:framework} further visualizes how different threats occur throughout a typical pipeline of deep graph learning. As a comparison, inherent noise or distribution shift typically happens in the data generation process due to sampling bias or environment noise without deliberate human design, while adversarial attacks are intentionally designed by malicious attackers after the data generation phase. Overall, we can inspect the three types of threats in a unified view from the uncertainty modeling framework. For a more in-depth discussion, see Section~\ref{sec:reliable_discusion}.

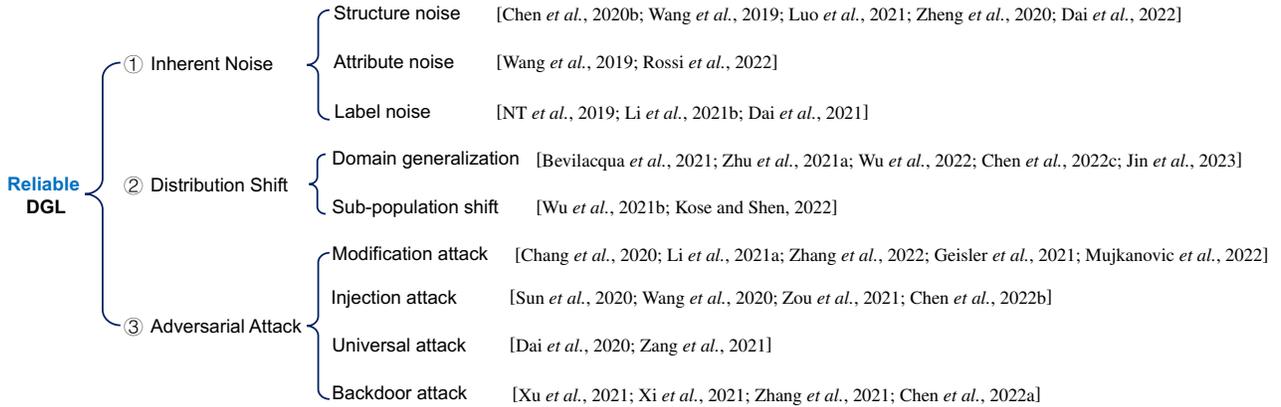
\begin{figure*}
    \centering
    \begin{picture}(500,160)
        \put(0,0){\includegraphics[width=\linewidth]{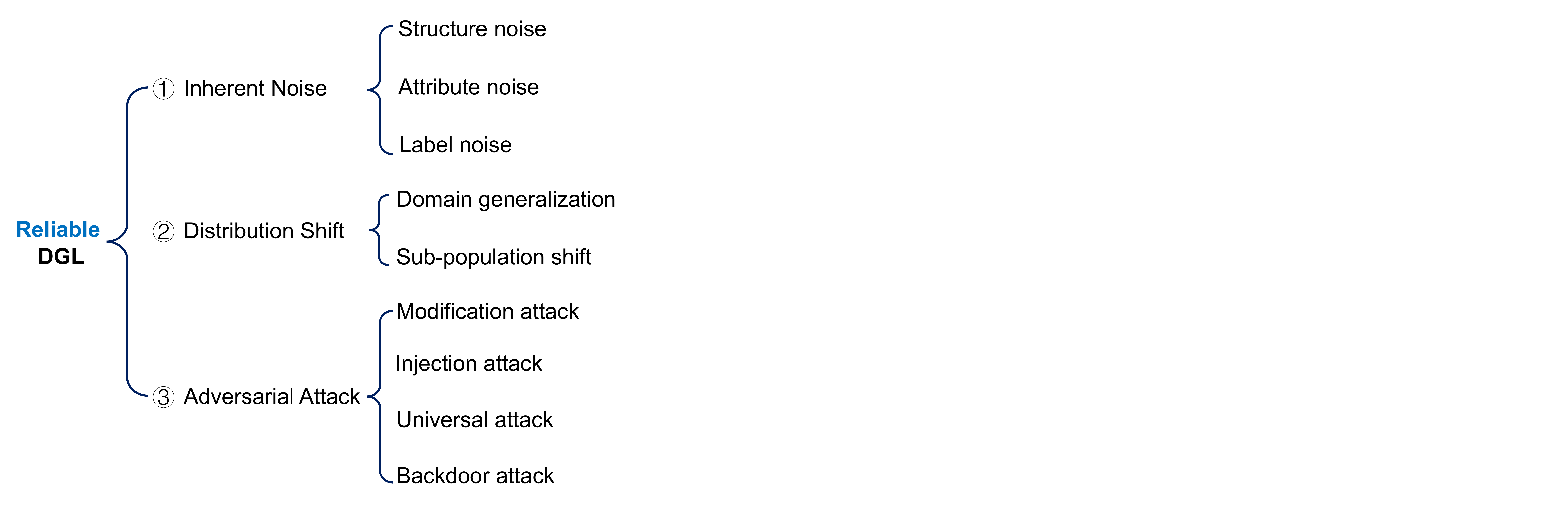}}
        \put(190,152){\scriptsize\cite{chen2020iterative,wang2019learning,luo2021learning,zheng2020robust,dai2022towards}}
        \put(190,134){\scriptsize\cite{wang2019learning,feature_propagation}}
        \put(190,115){\scriptsize\cite{nt2019learning,li2021unified,dai2021nrgnn}}
        \put(205,97){\scriptsize\cite{DBLP:conf/icml/BevilacquaZ021,zhu2021shiftrobust,wu2022towards,ciga,ttt}}
        \put(205,79){\scriptsize\cite{DBLP:journals/corr/abs-2106-11133,kose2022fair}}
        \put(197,61){\scriptsize\cite{DBLP:conf/aaai/ChangRXHZC0H20,li2021adversarial,zhang2022unsupervised,geisler2021robustness,mujkanovic2022are}}
        \put(195,45){\scriptsize\cite{DBLP:conf/www/SunWTHH20,DBLP:journals/datamine/WangLSLYZ20,DBLP:conf/kdd/ZouZDGKLT21,chen2022understanding}}
        \put(195,27){\scriptsize\cite{DBLP:journals/corr/abs-2011-14365,gua}}
        \put(196,8){\scriptsize\cite{DBLP:conf/wisec/XuXP21,DBLP:conf/uss/XiPJ021,DBLP:conf/sacmat/ZhangJWG21,DBLP:journals/corr/abs-2201-06202}}
    \end{picture}
    \caption{Overview of recent advances in reliable deep graph learning, classified into three groups according to mainstream reliability threats. Please see Sections \ref{sec:noise}, \ref{sec:dshift} \& \ref{sec:adv} for detailed introductions of them and corresponding enhancing techniques. Due to space limitation, here we only outlines the primary methods.}
    \label{fig:summary}
\end{figure*}

Over the past few years, numerous work has emerged to improve the reliability of DGL algorithms against the above threats from different perspectives, such as robust model architecture, optimization policy design, and uncertainty quantification.
In this survey, we explore recent advancements in this research direction. Specifically, this survey is organized as follows: we first review recent work from three aspects, namely inherent noise, distribution shift, and adversarial attack (a summarized list can be found in Figure~\ref{fig:summary}). For each aspect, we first summarize existing threats (e.g., various adversarial attacks). We then introduce typical reliability-enhancing techniques and discuss their limitations and future research directions. Lastly, we discuss the relationships among the above three aspects and describe some important open research problems yet to be addressed.

\nosection{Relation to existing surveys}
There are several surveys related to our paper. \cite{DBLP:journals/corr/abs-1812-10528} and \cite{DBLP:journals/corr/abs-2003-05730} both comprehensively review adversarial attacks on graphs.
\cite{DBLP:journals/corr/abs-2103-03036} focus on the robust representations of graph structure learning, and also briefly present an overview of the recent progress. These works mainly focus on the vulnerability of DGL methods against adversarial attacks. Given the large popularity of research on adversarial attacks, only two recent work \cite{DBLP:journals/corr/abs-2112-06070} and \cite{dshift_survey} focus on the reliability of current DGL algorithms against (structure) noise and distribution shift, respectively.
Our work stands apart from the aforementioned surveys in that it is not restricted to the scope of a single threat (e.g., adversarial attack). Instead, we systematically review the current advances and trends of reliable deep graph learning towards inherent noise, distribution shift, and adversarial attacks.

\section{Reliability against Inherent Noise}
\label{sec:noise}
Inherent noise refers to the noise that naturally exists in graph data.
It is unavoidable that real-world graph data would potentially include some inherent noises, which might stem from error-prone data measurement or collection, information loss during data preprocessing, sub-optimal graph structure w.r.t. downstream tasks, and so on. This section first discusses how different inherent noises can affect DGL algorithms, and then summarizes the recent typical techniques that can explicitly or implicitly alleviate the effects caused by inherent noises occurring in the pipeline of DGL.

\subsection{Inherent noise on graphs}
Inherent noises in graph data may exist in graph structure, node attributes, and corresponding node/graph labels, which draws out three types of inherent noises, i.e., structure noise, attribute noise, and label noise. Formally, we define a dataset $\mathcal{D}$ with inherent noise as:

\begin{equation}
    \mathcal{D}=(\mathbf{A}+\epsilon_a, \mathbf{X}+\epsilon_x, Y+\epsilon_y),
\end{equation}
where $\epsilon_a, \epsilon_x, \epsilon_y$ denote structure, attribute, and label noises respectively. We detail them as follows.

\nosection{Structure noise} The inherent structure noise may inevitably be introduced due to error-prone data measurement or collection \cite{chen2020iterative,wang2019learning}, 
e.g., protein-protein interaction networks
\cite{guo2021dockstream}. And it could also come from sub-optimal graph construction with task-irrelevant edges \cite{chen2020iterative,luo2021learning}, as the underlying motivation of establishing edges might not be always relevant to a specific downstream task \cite{zheng2020robust}.
\cite{dai2022towards} propose to learn a denoised and dense graph guided by the raw attributed graph to facilitate message passing for noise-resistant GNNs. Since most existing DGL algorithms rely on the graph topology to propagate information, the structure noise presents negative effects on the propagating process and consequently the final model performance.

\nosection{Attribute noise} There are two kinds of inherent attribute noise. On the one hand, the raw attributes of nodes may be incomplete or incorrect \cite{wang2019learning,feature_propagation}. For example, users may intentionally provide fake profiles due to privacy concerns in social networks. On the other hand, another subtle source is information loss when transforming raw node attributes into node embeddings, e.g., using the word embedding technique to encode textual data of nodes. During the training phase, the attribute noise in each node can be also directly passed to its neighbors, which would gradually affect the final node/graph embeddings.

\nosection{Label noise} The erroneous annotations in node or graph labels lead to the inherent label noise. Similarly to image classification, unavoidable erroneous annotations can occur in node/graph classification. Compared to images, it might be however more tricky to annotate labels for nodes due to structure dependency, which results in more noises or label sparsity issues \cite{li2021unified,nt2019learning}. Moreover, annotating labels for graphs becomes prohibitively expensive because of requiring more domain expertise. For instance, to annotate a drug-related property of a molecular graph, it needs to conduct experiments for measurement or extract the property from literature, which are both likely to introduce graph label noise \cite{ji2022drugood}. It is worth noting that, the previous study has shown that deep learning models tend to overfit label noise, which is also likely to happen in DGL models \cite{li2021unified,dai2021nrgnn}. 

\subsection{Techniques against inherent noise}
A number of methods have been presented for enhancing the robustness of DGL algorithms against above inherent noises. Although some of these methods are not originally designed for mitigating inherent noises, the key ideas behind these methods are useful for building robustness-enhancing techniques, and are thus also covered in this survey. In this subsection, we review these methods from data and regularization perspectives.

\nosection{Data denoising}
From the data perspective, data denoising is a straightforward way for reducing the noise effects in DGL algorithms.
For \emph{structure noise}, one natural idea is to assign learnable weights for each edge when perform node information aggregation in GNNs, so that the weight can hopefully reflect the tasks-desired graph structures. One of well-known techniques is incorporating self-attention modules into the original GNNs \cite{chen2020iterative}. As a more explicit way to denoise graph data, \cite{zheng2020robust} and \cite{luo2021learning} propose to prune task-irrelevant edges using a multi-layer neural network to draw a subgraph from a learned distribution.
The reparameterization trick \cite{DBLP:conf/iclr/JangGP17} is also applied to make this process differentiable, which can thus be jointly trained with its subsequent model for a specific task.
Regarding \emph{attribute noise}, \cite{feature_propagation} propose diffusion-based feature reconstruction to handle missing node attributes.
Instead of reconstruction incomplete attributes, to our best knowledge, there is currently no methods specially designed for denoising attribute noise, which can be a promising future direction.
For \emph{label noise}, \cite{nt2019learning} first observe label noise can lead to significant performance drops in the DGL setting. Motivated by prior methods for non-graph machine learning, this work directly applies the backward loss label correction
to train GNNs without considering graph structure. More recently, some work incorporates the structure information into the design. \cite{li2021unified} conduct sample reweighting and label correction by using random walks over graph structure for label aggregation to estimate node-level class probability distributions. \cite{dai2021nrgnn} propose to generate accurate pseudo labels, and assign high-quality edges between unlabeled nodes and (pseudo) labelled nodes to reduce label noise.

\nosection{Regularization}
From the regularization perspective, previous studies attempt to reduce overfitting to inherent noises by decreasing the complexity of the hypothesis space implicitly or explicitly.
On the one hand, various regularization techniques specially for GNNs are presented for reducing overfitting of DGL algorithms. The core idea behind these techniques is to randomly drop edges \cite{rong2019dropedge}, nodes \cite{hamilton2017inductive}, or hidden representations of GNNs \cite{chen2018fastgcn}. \cite{hasanzadeh2020bayesian} take a further step to unify these approaches into a Bayesian graph learning framework. Besides, an advanced data augmentation strategy namely Mixup is recently applied to DGL and is proved to be effective for several graph-related tasks~\cite{g_mixup}. On the other hand, there are some studies trying to explicitly impose regularization to the hypothesis space, i.e., building GNN models with some predefined constraints or inductive bias, which share a similar idea with approaches demonstrated in Section \ref{subsec:adv_enh}.
Moreover, from the Bayesian perspective, prior ensemble approach for DGL can also be seen as an implicit regularization, which independently trains multiple graph learning models with different input data transformations, then aggregates outputs from multiple models as the final output \cite{papp2021dropgnn} to mitigate the effects of noises. 

\nosection{Discussion}
Message passing scheme, although prevalently used in current DGL models, is a double-edged sword particularly on noisy data. Noise will propagate through the entire graphs so that representations of nodes may be negatively affected. Therefore, having a proper selection mechanism (e.g., attention) during propagation is promising yet less explored to facilitate message passing for robust DGL models.
In addition, it is interesting and probably more useful to simultaneously diminish multiple noises from structure, attributes, and labels, since denoising one type of noise often requires other type(s) information (desirably without noises) for assistance.

\section{Reliability against Distribution Shift}
\label{sec:dshift}
Distribution shift also arises naturally in many graph learning applications and presents severe challenges to current DGL methods.
Mathematically, it is a challenging situation where the distribution of datasets differs between the training and test stages, described as follows:
\begin{equation}
    \text{P}_{\text{train}}(\mathcal{G}, Y) \neq \text{P}_{\text{test}}(\mathcal{G}, Y),
\end{equation}
where $\text{P}(\mathcal{G}, Y)$ is the probability distribution of a graph dataset. Distribution shift on classic data formats (e.g. vision or texts) have been comprehensively investigated in some recent surveys (e.g., \cite{DBLP:conf/ijcai/0001LLOQ21,zhou2021domain}). 
However, there are inadequate discussions on graph data.
Here we provide a review of distribution shift literature with a focus on graph-structured data. This section first provides a categorization for typical distribution shift on graph data then introduces recent work for improving the reliability of DGL methods against distribution shift.

\subsection{Distribution shift on graphs}
Motivated by prior work focusing on distribution shift in general machine learning, we categorize distribution shift on graph data into two types, \emph{domain generalization} and \emph{sub-population shift}. Domain generalization refers to the training and testing distributions consist of distinct domains. Some typical examples include covariate shift and open-set recognition. While sub-population shift refers to training and testing distributions consist of the same group of domains but differ in the frequencies of each domain.

\nosection{Domain generalization}
One typical example of domain generalization on graph-related tasks is covariate shift, which assumes that the conditioned label distribution is the same for both training and test domains but differs in the data marginal distribution~\cite{DBLP:conf/icml/BevilacquaZ021}. For examples, in drug discovery, the scaffolds of drug molecules often differ at inference and in social networks and financial networks, the graph structures may significantly change with time~\cite{wu2022towards}.
Open-set recognition refers to the cases that new classes may appear at testing time, which commonly exists in e-commerce networks and recommendation systems where new users or items arrive from time to time.

\nosection{Sub-population shift}
Sub-population shift on graphs raises when the frequencies of data/label sub-populations change~\cite{kose2022fair}, which widely exists in many graph learning tasks, such as algorithmic fairness  and label shift.
Specifically, fairness issues of DGL could be caused by  societal bias contained in graph data~\cite{kose2022fair}.
Label shift refers to the cases where the marginal label distribution changes for two domains but the conditional distributions of the input given label stay the same across domains~\cite{DBLP:journals/corr/abs-2106-11133}. In addition, class-imbalance problem on graphs is a special case of label shift~\cite{DBLP:journals/corr/abs-2106-11133}. For instance, the label distribution is uniform for the testing distribution but is not for the training distribution.

\subsection{Techniques against distribution shift}
\label{sec:dshift_en}
There have been some methods proposed to tackle the challenges raised by distribution shift on graphs, which could mainly be classified into three categories: graph invariant learning, robust training, and uncertainty quantification.

\nosection{Graph invariant learning}
Graph invariant learning aims to learn invariant graph representations across different domains. The idea of invariant representation learning proposed recently has been adapted in several DGL models.
\cite{DBLP:conf/icml/BevilacquaZ021} use a causal model to learn approximately invariant graph representations that well extrapolate between the training and testing domains.
\cite{wu2022towards} inherit the spirit of invariant risk minimization 
to develop an explore-to-extrapolate risk minimization framework that facilitates GNNs to leverage invariant graph features for node-level prediction. 
\cite{zhu2021shiftrobust} adopt a distributional shift metric as a regularization to directly minimize the discrepancy between a biased trained datasets and unbiased IID dataset.
\cite{ciga} propose a contrastive framework to tackle graph domain generalization with various kinds of distribution shifts characterized by structural causal models.

\nosection{Robust training}
Robust training proposes to enhance the model robustness against distribution shift either by incorporating data augmentation or modifying the training framework. On the one hand, some methods generalize advanced augmentation techniques for general data format (e.g., mixup) to graph data. \cite{DBLP:journals/corr/abs-2106-11133} present a mixup-based framework for improving class-imbalanced node classification on graphs, which performs feature mixup on a constructed semantic relation space and edge mixup. \cite{kose2022fair} propose a fairness-aware data augmentation framework on node attributes and graph structure to reduce the intrinsic bias of the obtained node representation by GNNs. On the other hand, some methods propose to integrate adversarial training techniques in DGL models. \cite{DBLP:journals/corr/abs-2010-09891} present a method that iteratively augments node features with adversarial perturbations during training and helps the model generalize to out-of-distribution (OOD) samples by making it invariant to small fluctuations in input data.

\nosection{Uncertainty quantification}
Apart from the above two directions that aim to improve model robustness, uncertainty quantification can be seen as a complementary way to enhance the reliability of DGL algorithms, since the estimated uncertainty can be used for rejecting unreliable decisions with high model predictive uncertainties. Here we introduce some recent work of uncertainty quantification for DGL algorithms.
A natural uncertainty measure can be the prediction confidence, i.e., the maximum value of the Softmax output. However, recent work observes that GNNs with Softmax prediction layer are typically under-confident, thus the confidence cannot precisely reflect the predictive uncertainty \cite{wang2021confident}. There are two ways to solve this problem.
First, recent work introduces probabilistic blocks into original GNNs for modeling the posterior weight distribution, which can provide more accurate uncertainty estimation than deterministic GNN architectures. For example, \cite{han2021reliable} propose to replace the Softmax decision layer with a Gaussian process block, which provides accurate uncertainty estimations. Unlike this work, Bayesian GNNs \cite{hasanzadeh2020bayesian} aggressively transform whole different layers into Bayesian counterparts. Concretely, it treats both model weights and sampling process as probabilistic distributions and adopts variational inference to estimate the parameters of these distributions.
Second, a more straightforward way is to perform confidence calibration in a post-hoc fashion without modifying the GNN architectures. One typical calibration method is temperature scaling. However, it is originally designed for DNNs and is proved to have poor performance in the DGL setting. \cite{wang2021confident} adapt temperature scaling to GNNs by employing additional GNNs to predict a unique temperature for each node. Since temperatures are produced by considering both node attributes and graph topology, this method achieves better calibration performance compared to the original method.

\nosection{Discussion}
Most efforts have been made on tackling the distribution shift problems from
modeling or training perspectives, which requires additional cost of modifying model architectures or optimizing model parameters. These limitations motivate a new line of research that focused on graph test-time training. Graph test-time training aims to adapt models based on test samples in the presence of distributional shifts, which allows better robustness at inference time by solving a test-time task. A recent attempt~\cite{ttt} proposes a graph transformation framework to refine graph data at test time, yielding state-of-the-art performance in different challenging situations including distribution shift. In conclusion, graph test-time training is a valuable direction toward more robust DGL models on OOD graphs.

\section{Reliability against Adversarial Attack}
\label{sec:adv}
Adversarial attacks aim to cause a model to make mistakes with carefully-crafted unnoticeable perturbations (adversarial samples) or predefined patterns (backdoor triggers).
The goal of adversarial attacks is to maximize the loss of DGL models with imperceptible perturbations on the original graph, which we define with:
\begin{equation}
    \text{Find} \ \hat{\mathcal{G}} \quad \text{s.t.}\quad \hat{\mathcal{G}}= \arg \max_{\hat{\mathcal{G}} \approx \mathcal{G}} \mathcal{L}(\hat{\mathcal{G}}, Y),
\end{equation}
where $\hat{\mathcal{G}}$ is the perturbed graph with some imperceptible constraints.
Literature has revealed that state-of-the-art DGL algorithms remain highly vulnerable to adversarial samples, posing significant security risks to several application domains \cite{DBLP:journals/corr/abs-1812-10528}.
Yet, the adversarial reliability of GNNs is highly desired for many real-world systems, especially in security-critical fields. In this subsection, we provide an overview of adversarial attacks on graphs and subsequently review recent works that mitigate such threats.

\subsection{Adversarial attack on graphs}
With rising concerns addressed on the reliability of DGL models, there has been an explosion of papers around attacking such models and finding their vulnerabilities. Literature has categorized adversarial attacks into several typical dimensions \cite{DBLP:journals/corr/abs-1812-10528,DBLP:journals/corr/abs-2003-05730}. Specifically, adversarial attacks can be performed in the training phase (poisoning attack) and the inference phase (evasion attack), to mislead the prediction of the model on specific important instances such as nodes (targeted attack), or degrade the overall performance of the model (non-targeted attack).
Based on the way employed by attackers to perturb the graph data or learning model, this survey reviews prior work from four dimensions: \emph{manipulation attacks}, \emph{injection attacks}, \emph{universal attacks}, and \emph{backdoor attacks}. Specifically, manipulation, injection attacks, and universal attacks can be performed in the inference phase while the backdoor attacks always happen in the training phase.

\nosection{Manipulation attacks}
In manipulation attacks, attackers generate adversarial samples by modifying either the graph structure or node attributes. For instance, attackers can add, remove, or rewire an edge in the graph to generate legitimate perturbations. 
As a pioneering work, \cite{DBLP:conf/kdd/ZugnerAG18} craft adversarial samples by manipulating both edges and attributes of the graph with a greedy search algorithm. Followed by this work, the gradient-based approach becomes a prominent way to craft adversarial samples. By exploiting the gradient information of the victim model or a locally trained surrogate model, attackers can easily approximate the worst-case perturbations to perform attacks
\cite{li2021adversarial,XuC0CWHL19,Wu0TDLZ19}. While current research heavily relies on supervised signals such as labels to guide the attacks and are targeted at certain downstream tasks, \cite{zhang2022unsupervised} propose an unsupervised adversarial attack where gradients are calculated based on graph contrastive loss. In addition to gradient information, \cite{DBLP:conf/aaai/ChangRXHZC0H20}
approximate the graph spectrum to perform attacks in a black-box fashion.

\nosection{Injection attacks}
The manipulation attack requires the attacker to have a high privilege to modify the original graph, which is impractical for many real-world scenarios. Alternatively, injection attacks has recently emerged as a more practical way that injects a few malicious nodes into the graph. The goal of the injection attack is to spread malicious information to the proper nodes along with the graph structure by several injected nodes. In the poisoning attack setting, \cite{DBLP:conf/www/SunWTHH20} first study the node injection attack and propose a reinforcement learning based framework to poison the graph. \cite{DBLP:journals/datamine/WangLSLYZ20} further derive an approximate closed-from solution to linearize the attack model and inject new vicious nodes efficiently. In the evasion attack setting, \cite{DBLP:conf/kdd/ZouZDGKLT21}
consider the attributes and structure of injected nodes simultaneously to achieves better attack performance. However, a recent work \cite{chen2022understanding} shows that the success of injection attacks is built upon the severe damage to the homophily of the original graph. Therefore, the authors present to optimize a harmonious adversarial objective to preserve the homophily of graphs.

\nosection{Universal attacks} Both manipulation and injection attacks on graphs are designed for a target-dependent scenario. Differently, the universal attack crafts a single and unique perturbation that is capable to fool a GNN when applied to any target node. Such attacks incur a lower cost for attacks as the perturbations are generated once and for all. \cite{gua} first study the universal attacks on graphs, which achieves the adversarial goal by flipping edges connected to a set of anchor nodes. In the follow-up work, the attack capability is enhanced through a small number of malicious nodes connected to them \cite{DBLP:journals/corr/abs-2011-14365}. Although universal attacks in vision research have been extensively researched,
there is still significant room for exploration in the graph domain.

\nosection{Backdoor attacks}
In contrast to above attacks, backdoor attacks aim to poison the learned model by injecting \emph{backdoor triggers} into the graph at the training stage. As a consequence, a backdoored model would produce attacker-desired behaviors on trigger-embedded inputs (e.g., misclassify a node as an attacker-chosen target label) while performing normally on other benign inputs. Typically, a backdoor trigger can be a node or a (sub)graph designed by attackers.  \cite{DBLP:conf/uss/XiPJ021} and \cite{DBLP:conf/sacmat/ZhangJWG21} propose to use subgraphs as trigger patterns to launch backdoor attacks. \cite{DBLP:conf/wisec/XuXP21} select the optimal trigger for GNNs based on explainability approaches. Backdoor attacks are a relatively unexplored threat in the literature. However, they are more realistic and dangerous in many security-critical domains as the backdoor trigger is hard to notice even by human beings.
The above attacks forge a backdoored GNN by perturbing its model parameters with poisoned inputs (poisoning attacks), which would result in a performance degradation even triggers are not activated.
To this end, \cite{DBLP:journals/corr/abs-2201-06202} propose a fast and effective approach that generates a trigger node without modifying the model parameters.

\subsection{Techniques against adversarial attack}
\label{subsec:adv_enh}
While there are numerous (heuristic) approaches aimed at robustifying GNNs, there is always a newly devised stronger attack attempts to break them, turning into a veritable arms race between attackers and defenders. To avoid endless arms races with attackers, extensive efforts have been made to mitigate adversarial attacks in different ways.
In this subsection, we review recent robustness-enhancing techniques of DGL against adversarial attacks from data, model, optimization, and certification perspectives.

\nosection{Graph processing}
From data perspective, a natural idea is to process the training/testing graph to remove adversarial perturbations thus mitigating the negative effects.
Currently, enhancing approaches in this direction are mainly supported by empirical observations on specific attacks. For example, there is a tendency of adversarial attacks to add edges between nodes with different labels and low attribute similarity. Motivated by such an observation, \cite{Wu0TDLZ19} prune the perturbed edges based on the Jaccard similarity of node attributes, with the assumption that adversarially perturbed nodes have low similarity to most of their neighbors.
Further, the clean graph structure can be learned simultaneously by preserving graph properties of sparsity, low rank, and feature smoothness during training~\cite{jin2020graph}.
Graph processing based methods are cheap to implement while significantly improving the adversarial robustness of GNNs. However, empirical observation based on specific attacks makes such methods difficult to resist unseen attacks.

\nosection{Model robustification}
Refining the model to prepare itself against potential adversarial threats is a prominent enhancing technique and we term it as model robustification. Specifically, the robustification of GNNs can be achieved by improving the \emph{model architecture} or \emph{aggregation scheme}. There are several efforts that aim to improve the architecture by employing different regularizations or constraints on the model itself, such as 1-Lipschitz constraint \cite{DBLP:conf/icml/ZhaoZZWJZJD021}, $\ell_1$-based graph smoothing \cite{DBLP:conf/icml/LiuJ0LLW0T21} and adaptive residual \cite{liu2021graph}. As recently shown in \cite{geisler2021robustness,chen2021understanding}, the way that GNNs aggregate neighborhood information for representation learning makes them vulnerable to adversarial attacks. To address this issue, they derive a robust median function instead of a mean function to improve the aggregation scheme. Overall, a robustified model is resistant to adversarial attacks without compromising on performance in benign situations.

\nosection{Robust training}
Another enhancing technique that successfully applied to the GNN model is based on the robust training paradigms. Adversarial training is a widely used practical solution to resist adversarial attacks, which builds models on a training set augmented with handcrafted adversarial samples.
Essentially, the adversarial samples can be crafted via specific perturbations on the graph structure~\cite{lisat} or node attributes~\cite{DBLP:journals/tkde/FengHTC21}.
Although adversarial training can improve the generalization capability and robustness of a model against unseen attacks, it inevitably introduces additional overheads during training and suffers from overfitting on adversarial samples.

\nosection{Robustness certification}
Robustness certification is a measure to verify whether a single instance (e.g., a node or graph) is certifiably robust under worst-case adversarial perturbations.
It typically provides a lower bound on the actual robustness while attacks provide an upper bound~\cite{mujkanovic2022are}.
Robustness certification was initially proposed as an assessment method. Over the past few years, it has been adopted as a guideline to improve robustness against adversarial attacks.
A few recent works study certified robustness of GNNs against adversarial perturbations on node attributes~\cite{DBLP:conf/kdd/ZugnerG19} or graph structure~\cite{DBLP:conf/kdd/WangJCG21}.
Different from previous robustness certificates which independently consider perturbation of each prediction, \cite{schuchardt2021collective} consider collective robustness certificate that computes multiple predictions which are simultaneously guaranteed to remain stable under perturbations.
Although certified robustness can guide the defenders for more reliable architecture design, it is relatively less explored and has received little attention from the research community.

\nosection{Discussion}
The adversarial robustness is highly desirable for a model being trust in real-world applications but the evaluation of robust DGL models against adaptive attacks is rarely explored.
However, current defenses are usually heuristic and only effective for certain attacks rather than all attacks.
Until recently, \cite{mujkanovic2022are} have shown that the adversarial robustness of defenses is overestimated and they can be easily broken under adaptive attacks. To enable DGL models to predict robustly in reality, it is therefore crucial to evaluate DGL models with more sophisticated attacks.

\section{Overall Discussion}
\label{sec:reliable_discusion}
Given the above comprehensive summary of recent advances in reliable DGL research, we further provide overall discussions among the above topics including their relations, differences, and applications.

\nosection{A unified view}
The uncertainty modeling framework allows us to examine the three types of threats in a unified manner.
There are two typical types of uncertainties, \emph{aleatoric} and \emph{epistemic} uncertainties.
Specifically, we can treat inherent noise as the main source of aleatoric uncertainty, since these noises are irreducible even given access to infinite samples.
In addition, we can treat adversarial attack and distribution shift as two sources of epistemic uncertainty, as we can reduce these uncertainties by introducing data samples on different domains/sub-populations through advancing data augmentation.
For example, to combat with adversarial noises, robust training~\cite{XuC0CWHL19} involves adversarial samples into the training process to enhance the adversarial robustness of GNN models.
Under this unified view, we can leverage prior uncertainty estimation methods to detect adversarial samples (adversarial attack), out-of-distribution samples (distribution shift), and outliers (inherent noises), which provides further information for enhancing model's reliability.

\nosection{Difference among above threats}
In general, the above three types of threats can all be seen as the ``mismatch'' between training and testing samples. Here, we highlight subtle differences among them. The inherent noise or distribution shift happens in the training data generation process due to sampling bias and environment noise without deliberate human design, while adversarial attacks are deliberately designed by malicious attackers after the training/testing sample generation phase. Furthermore, the inherent noise is typically irreducible, while the other two types of threats can be alleviated by sampling more data with expertise. Despite the differences, some general techniques can be used for preventing the three types of threats. For example, one can inject probabilistic decision module into GNNs to provide predictions and its uncertainty estimation \cite{han2021reliable,hasanzadeh2020bayesian}.
Uncertainty estimation can be further enhanced by introducing OOD and adversarial samples into the training phase. Thus, the estimation can be used to detect the above threats, improving the decision reliability of various DGL algorithms.

\nosection{Difference to general reliable machine learning}
Tremendous efforts have been made to improve the reliability of deep learning on non-graph data such as images and texts. In contrast to these methods, improving the reliability of GNNs on graph data poses unique challenges.
Due to message-passing mechanism used by most DGL algorithms,
the adversarial perturbation, inherent noise and distribution shift of one node can be transmitted to its neighbors and further hinder the model performance. Therefore, previous general reliable machine learning algorithms need to be modified by considering the relationship between different nodes.

\section{Conclusion and Future  Directions}

This survey gives an overview of recent advances in researching the reliability of DGL algorithms in terms of three fundamental aspects: inherent noise, distribution shift, and adversarial attack.
For each threat, we provide a systematical view to inspect advancing robustness-enhancing techniques for DGL.
We hope this survey can help researchers to better understand relations between different threats and to choose appropriate techniques for their applications.
Finally, we summary several challenges and opportunities worthy of future explorations.

\nosection{Theoretical framework}
Despite algorithmic advances for reliable DGL, there is still a lack of theoretical frameworks to formally analyze the effectiveness of these methods. For example, how to analyze out-of-distribution generalization bound in the DGL setting remains an open problem.

\nosection{Unified solution}
Section~\ref{sec:reliable_discusion} shows relations among three threats. In a real-world setting, these threats may happen simultaneously. Therefore a unified solution to mitigate the effects caused by these threats is desired.

\nosection{Connection to learning stability}
As pointed out by prior work for general ML, there is a strong connection between robustness and learning stability. Thus, from the optimization perspective, how to build robust learning algorithms for DGL is also an interesting direction. Prior work~\cite{wu2022towards} mentioned in Section~\ref{sec:dshift_en} can be seen as an initial attempt, which applies invariant risk minimization to DGL settings to learn a stable graph representation.

\nosection{Scalability and adversarial robustness}
Existing studies of the reliability to adversarial attacks mainly focus on relatively small graphs (e.g., 2-3k nodes), which suffer from the scalability issue on real-world large-scale and dynamically evolving graphs.
In this respect, highly scalable and robust DGL algorithms remain largely unexplored, while there is only a recent work~\cite{geisler2021robustness} attempting to address the challenge.

\nosection{Fairness and adversarial robustness}
Existing work prefers to utilize standard performance metrics, such as accuracy, to measure the robustness of a DGL model,
which is, however, apparently insufficient for evaluating the overall reliability performance.
However, such a metric is apparently insufficient for the evaluation of reliability from a single dimension.
A DGL algorithm with high accuracy may not be fair to different attribute groups, which results in severe disparities in accuracy and robustness between different groups of data.
This calls for future work to explore fair and robust DGL algorithms and develop principled evaluation metrics beyond accuracy.

\nosection{Counterfactual explanations \& adversarial samples}
One major problem of DGL algorithms is the lack of interpretability. To address this issue, counterfactual explanations are proposed as a powerful means for understanding how decisions are made by algorithms. While prior research in vision tasks~\cite{DBLP:journals/corr/abs-2106-09992} has shown that counterfactual explanations and adversarial samples are strongly related approaches with many similarities, there is currently little work on systematically exploring their connections in DGL.
Therefore, the study of counterfactual explanations and adversarial samples is another promising research direction.

\nosection{Reliability benchmarks}
Along with the fast development of reliable DGL algorithms, real-world benchmarks are desired for facilitating the research community.
Some early benchmarks for specific settings have been established, e.g., \cite{ji2022drugood} present an OOD dataset curator and benchmark for AI-aided drug discovery designed specifically for the distribution shift problem with data noise. It is challenging yet promising to build a general benchmark platform to cover more reliability aspects mentioned in this survey.

\clearpage

\bibliographystyle{named}
\small
\bibliography{ijcai22}
\end{document}